\documentclass[11pt]{article}
\usepackage{amsmath}
\usepackage{graphicx}
\usepackage{amssymb}
\usepackage{xcolor}
\usepackage{microtype}
\usepackage{svg}
\usepackage[T1]{fontenc}
\usepackage[font=sf,labelfont={sf}]{caption}
\usepackage{sectsty}
\usepackage{mathtools}
\usepackage{titlesec}
\usepackage{soul}

\titlespacing*{\section}{0pt}{*1}{0pt}
\titlespacing*{\subsection}{0pt}{*1}{3pt}

\usepackage{natbib}

\usepackage[margin=1in]{geometry}
\allsectionsfont{\sffamily}

\usepackage{setspace}
\usepackage{lmodern}

\title{
Spatial Spiking Neural Networks Enable Efficient and Robust Temporal Computation
}
\author{
Lennart P. L. Landsmeer$^{1,2,*,}$\footnotemark[2],
Amirreza Movahedin$^{1,2,*}$,
Mario Negrello$^{2,1}$,
\\
Said Hamdioui$^1$,
Christos Strydis$^{2,1,}$\footnotemark[3]
\\
\\
\small
$^1$ Department of Quantum and Computer Engineering,
\\ \small
Delft University of Technology,
\small
Mekelweg 4, 2628 CD Delft, NL
\\
\small
$^2$ NeuroComputingLab, Department of Neuroscience,\\
\small
Erasmus Medical Center,
Dr. Molewaterplein 40, 3015 GD Rotterdam, NL\\
\\
\small
$^*$These authors contributed equally.\\
\small
{\footnotemark[2]  \hspace{0.1em} l.p.l.landsmeer@tudelft.nl}
\hspace{3em}
{\footnotemark[3] \hspace{0.1em} c.strydis@erasmusmc.nl}
}
\date{\small December 2025}

\begin{document}

 \fontfamily{lmss}\selectfont

\maketitle
\begin{abstract}
The efficiency of modern machine intelligence is measured by a high model accuracy and a low computational effort. Specifically in spiking neural networks (SNNs), synaptic delays are essential for capturing temporal patterns, yet current approaches treat delays as fully trainable, unconstrained parameters which incur large memory footprints, increased computational cost, and a disconnect from biological reality. In contrast, delays in the biological brain arise from the physical distance between neurons embedded in space. Inspired by this principle, we introduce Spatial Spiking Neural Networks (SpSNNs), a general framework in which neurons learn positions in a finite-dimensional Euclidean space and communication delays emerge from inter-neuron distances. This replaces per-synapse delay learning with position learning, drastically reducing the number of trainable parameters while preserving temporal expressiveness. Across two benchmark tasks (Yin-Yang and Spiking Heidelberg Digits), SpSNNs outperform conventional SNNs with unconstrained delays despite using far fewer parameters. Contrary to intuition, we find that accuracy consistently peaks in 2D and 3D networks rather than in infinite-dimensional ones, revealing a beneficial regularization effect. On top of that, dynamically sparsified SpSNNs retain their accuracy even at 90\% sparsity, achieving equal performance with up to 18$\times$ fewer parameters compared to standard, delay-trained SNNs. Given these outcomes and the fact that the spatial structure of SpSNNs can directly translate into geometrical placements, SpSNNs in principle lend themselves also to efficient hardware implementations. In terms of methodology, SpSNNs rely on calculating exact gradients for trainable delays by elegant automatic differentiation with novel, custom-derived gradients, thus allowing arbitrary neural models and network architectures. Thereby, our proposed SpSNNs provide a platform for exploring the spatial dimension of temporal expressivity in bio-inspired AI models and provide a hardware-friendly substrate for scalable, energy-efficient neuromorphic machine intelligence.
\end{abstract}


\section*{Introduction}

\begin{figure*}[t]
    \centering
    \includegraphics[width=\textwidth]{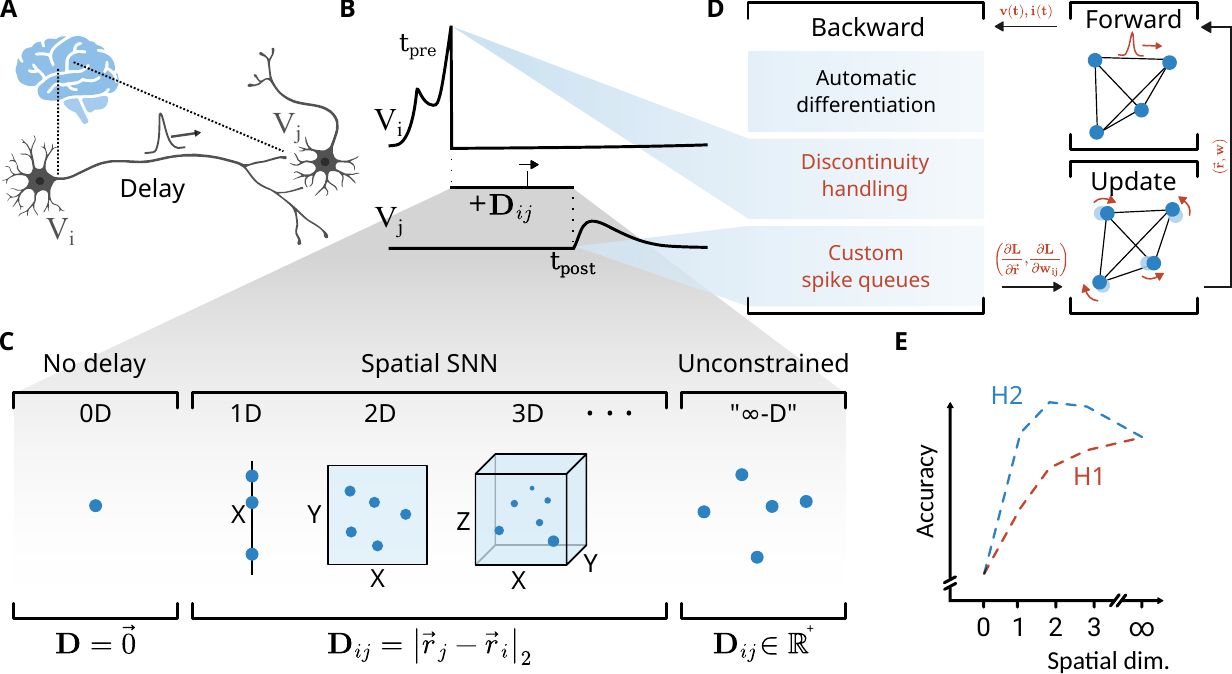}
    \caption{Spatial SNNs.
    \textbf{A:} Delays in the brain originate from the spatial distance a spike has to travel between neurons.
    \textbf{B:}
    In simulation, a delay is the difference between the presynaptic spike time $t_{\rm pre}$ and the postsynaptic spike time $t_{\rm post}$
    \textbf{C:}
    Existing implementations either use no delays or unconstrained trainable delays. This work introduced SpSNNs, where the delays are derived from the spatial distance between neurons.
    \textbf{D:}
    In our approach, exact gradients are calculated via modified automatic differentiation.
    The training cycle consists of a forward, a backward and an update phase.
    The forward pass is a time-discretized simulation, amended to carefully handle spike-induced updates. The backward pass, in contrast to existing technique, relies on automatic differentiation, with spike-induced updates handled by custom gradients (as derived in Methods). This makes swapping out different SpSNN architectures and neural models easy. Calculated loss-parameter gradients are finally used to update positions and weights.
    \textbf{E:} Two hypotheses for SpSNNs, knowing that adding trainable synaptic delay improves the accuracy of SNNs. Either constraining the dimensionality of the network decreases the accuracy (H1), or it improves the performance of the network (H2).
    }
    \label{fig:setup}
\end{figure*}

Biological neural networks excel at fast, energy-efficient computation in environments where information is encoded temporally~\cite{sengupta2013information,niven2016neuronal,meoier2017brainenergy}. As artificial intelligence (AI) increasingly confronts real-world settings that demand low-power, event-driven and temporally precise processing -- from robotics to neuromorphic hardware -- there is a growing need for models that capture the computational principles underlying such efficiency. Spiking neural networks (SNNs), which encode information in the timing of discrete spike events, offer a pathway toward this goal~\cite{maass1997snn}.  Recently, networks that incorporate trainable synaptic delays have been considered, and showed  that trainable delays are the key requirement to further improve the temporal performance of SNNs~\cite{hammouamri2023learningdelays,goltz2025delgrad,sun2024delay,sun2023learnable}. However, these trainable synaptic delays are unconstrained in the set of values that they can have.\\

In the brain, neurons populate a 3-dimensional space, and the inter-neuron communication delay is a result of their physical distance (Figures~\ref{fig:setup}A and~\ref{fig:setup}B). As a result, the value-set of these communication delays is constrained by geometry. Hence, according to this view of delays being a function of neuron positions, networks with unconstrained, trainable delays see neurons in an infinite-dimensional space. By the same token, networks with no trainable synaptic delays have a 0-dimensional neuronal space (Figure~\ref{fig:setup}C). Following this chain of thought, and while being inspired by observing the brain, in this work we bring the same geometrical constraints into SNNs. The hope is that, by restricting the originally infinite neuronal space to a finite-dimension one, the number of trainable parameters will decrease, thus resulting to more computationally efficient networks albeit at the potential cost of accuracy (Figure~\ref{fig:setup}E hypotheses H1 and H2).

To test these hypotheses, we capture the dimensionality of SNNs via Spatial Spiking Neural Networks (SpSNN). We define SpSNNs as a class of models whereby neurons are embedded in a Euclidean space with learnable positions and, thus, spike-propagation delays emerge from inter-neuron distances, as opposed to being unconstrained parameters (see Methods). In that way, training -- then -- optimizes synaptic weights and neuron coordinates jointly, resulting in SNNs with reduced parameter counts. When lowering dimensionality, our experiments reveal SpSNNs of lower complexity, and contrary to intuition, they also reveal an increase in accuracy for two- and three-dimensional SpSNNs.

By grounding temporal computation in learnable spatial structures, SpSNNs offer a biologically inspired and theoretically principled foundation for efficient spiking-based machine intelligence. Beyond their conceptual contributions, in this work we show that SpSNNs also confer practical advantages for deployment. Representing delay structure through low-dimensional position embeddings rather than large delay matrices reduces memory footprint and computational overhead, enabling more efficient execution on conventional hardware platforms such as CPUs, GPUs and AI accelerators~\cite{ landsmeer2024tricking}. Moreover, because learned neuron coordinates correspond directly to geometric placement, SpSNNs naturally map onto two- or three-dimensional neuromorphic chips, simplifying placement, routing and delay management in hardware implementations. These properties position SpSNNs as a
promising
substrate for scalable, energy-efficient spiking computation across both standard AI infrastructure and emerging neuromorphic systems.

\section*{Results}

\begin{figure*}[t]
    \centering
    \includegraphics[width=\textwidth]{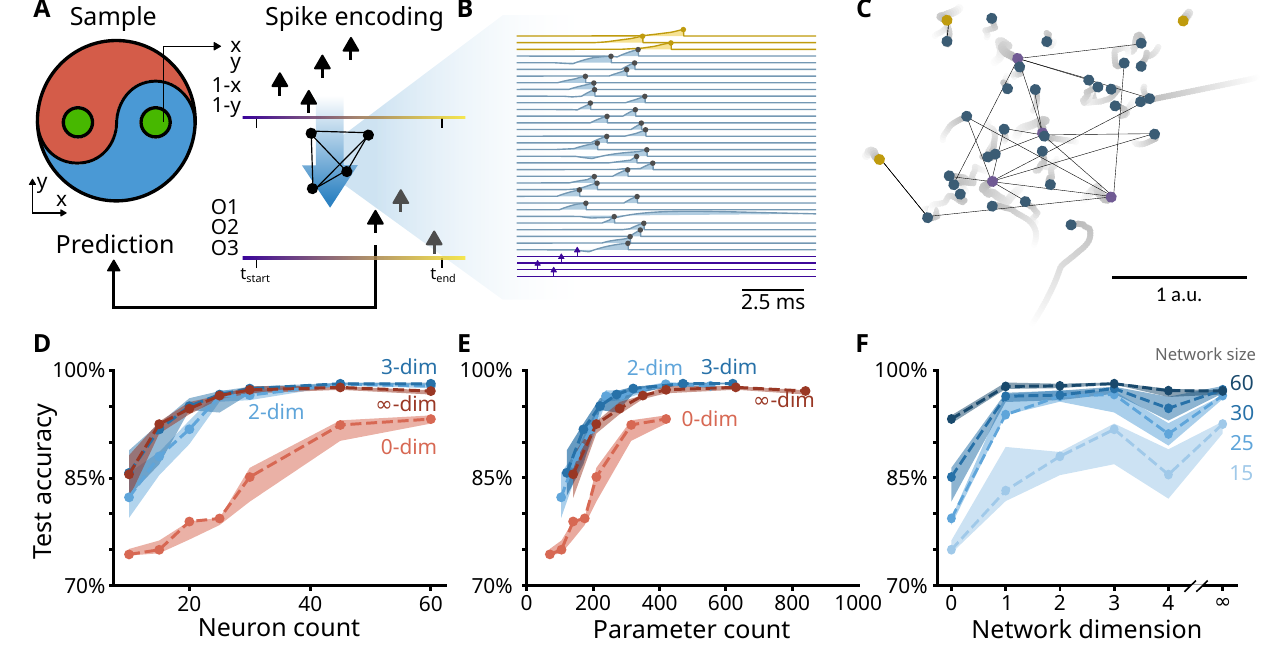}
    \caption{The YY Task.
    \textbf{A:} The task is classifying a point in the Yin-Yang plane. The input encoding is shown for a sample point. The output of the network is encoded with TTFS.
    \textbf{B:} The response of the network for an example input. Note that the SpSNNs for this task are structured in a feed-forward network, with each neuron being allowed to only spike once.
    \textbf{C:} Neurons' positions are optimized during the training process. Here shows the movement of neurons in a two-dimensional space throughout the training, in addition to the strongest connection of the network.
    \textbf{D:} Test accuracy versus the number of hidden neurons for SpSNNs of different dimensionalities. 0-dim and $\infty$-dim refer to only trainable weight and both trainable weight and delays respectively. The reported numbers are the median and interquartile ranges of 5 runs with different seeds. 
    \textbf{E:} Test accuracy versus the total number of trainable parameter. We can see that the two-dimensional SpSNN outperforms other dimensionalities, in addition to the $\infty$-dimensional network. 
    \textbf{F:} Test accuracy versus the dimension of the network for SpSNNs with different hidden neurons. Note the peak in almost all of the SpSNNs at two-dimensions.
    }
    \label{fig:yy}
\end{figure*}

\begin{figure*}[t]
    \centering
    \includegraphics[width=\textwidth]{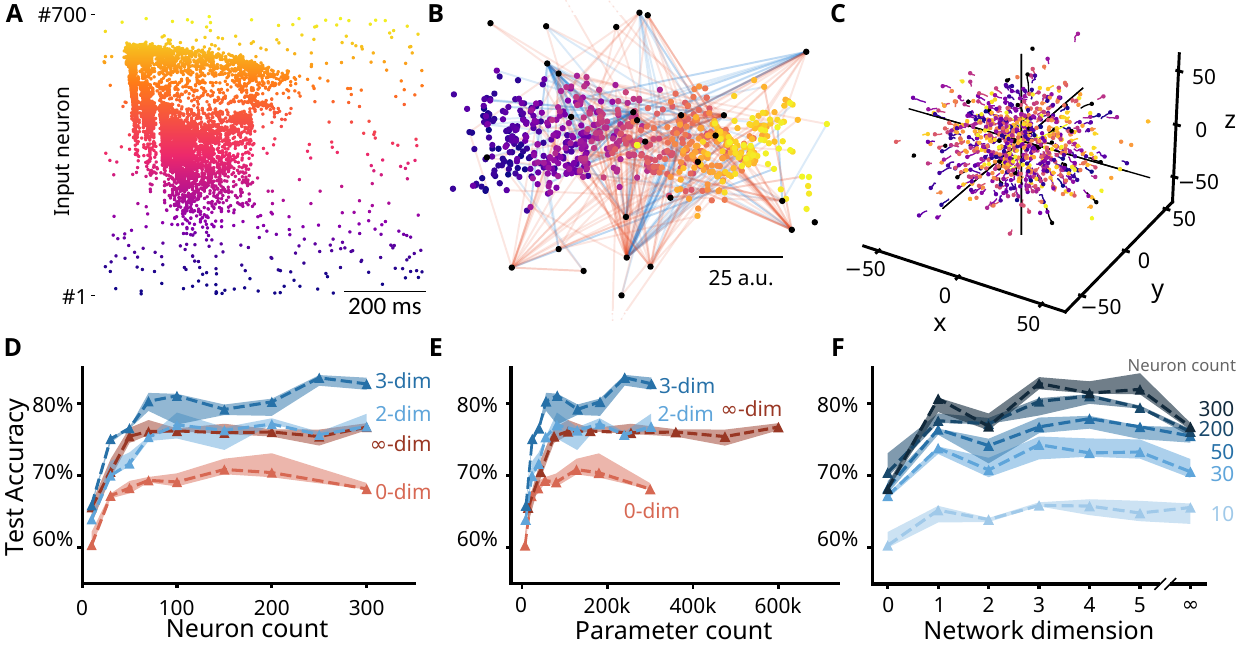}
    \caption{The SHD Task.
    \textbf{A:} SHD spoken digit example. Low index neuron represent low-frequency components, while high index neurons represent high-frequency-content.
    \textbf{B:} Two-dimensional SpSNN with 30 hidden neurons, trained on the SHD task. Input neurons are initialized on a line. Black points correspond to the learned hidden neuron positions. The rest of the points correspond to the input color in A. Blue and red lines correspond to connections with respectively synaptic weights larger than 1.5 or smaller than -1.5.
    \textbf{C:} The movement of neurons throughout the training of the three-dimensional SpSNN. The final learned positions are shown with the points. All positions are randomly initialized.
    \textbf{D:} Test accuracy versus the number of hidden neurons for SpSNNs of different dimensionalities. 0-dim and $\infty$-dim refer to only trainable weight and both trainable weight and delays respectively. The reported numbers are the median and interquartile ranges of 5 runs with different seeds.
    \textbf{E:} Test accuracy versus the total number of trainable parameter. We can see that the three-dimensional SpSNN outperforms other dimensionalities, in addition to the $\infty$-dimensional network. 
    \textbf{F:} Test accuracy versus the dimension of the network for SpSNNs with different hidden neurons. Note the peak in almost all of the SpSNNs at around three-dimensions.
    }
    \label{fig:shd}
\end{figure*}

\subsection*{SpSNNs can learn complex temporal patterns}

In order to capture the spatial essence of SNNs, we created SpSNNs that place neurons in a Euclidean space. Each neuron is assigned a position within that space, and the communication delay between two neurons is calculated based on their Euclidean distance. During the learning process, the positions of all the neurons alongside the synaptic weights are optimized. In this spatial view of SNNs, networks with unconstrained, trainable delays can be seen as infinite-dimensional and networks with no trainable delays as zero-dimensional.

To study the validity, performance, and generality of SpSNNs, and to test our hypotheses about the effect of dimensionality on accuracy (Figure~\ref{fig:setup}E), we looked at two diverse neuromorphic tasks: 1) Yin-Yang (YY)~\cite{kriener2022yin} is a simple, non-linear classification task for early benchmarking of SNNs. As input, the network receives coordinates of a point on a map and has to classify the point belonging to either Yin (red), Yang (blue), or Dot (green) region (Figure~\ref{fig:yy}A). 2) Spiking Heidelberg Digits (SHD)~\cite{cramer2020heidelberg} task is a collection of spoken digits in English and German languages (Figure~\ref{fig:shd}A). Therefore, the network's task is to classify the audio-based input spikes into 20 different available classes. This is a more complicated and demanding task compared to YY, chosen to challenge our proposed methods. We chose these two tasks as our testbed since they are widely used, standardized tasks, specifically designed to benchmark SNNs~\cite{goltz2025delgrad, cramer2020heidelberg, Mészáros2025efficient}.

We configured SpSNNs with different network architectures, number of neurons, and information encoding schemes to accommodate to each of the tasks' needs. For both tasks, we examined the effect of different numbers of hidden neurons in addition to different dimensionalities of the space. For the YY task, we used Leaky Integrate-and-Fire (LIF) neurons~\cite{Gerstner2002spikingneurons} that could only spike once. The neurons are structured as a single-layer, feed-forward SpSNN where all input, hidden, and output neurons can adjust their positions during training. For the input neurons, a larger value is encoded with a later spike within the input spike window, and for the output neurons, a time-to-first-spike (TTFS) encoding is used, where the earliest arrival of a spike determines the network's classification output. For the SHD task, we organized the LIF-neurons in a recurrent SpSNN fashion with a linear readout. The network uses a rate-coded scheme, where a larger value is encoded with a higher frequency of spiking.

Our experiments reveal that SpSNNs learn both tasks by optimizing the positions of the neurons in different dimensions (Figures~\ref{fig:yy}C and~\ref{fig:shd}C). Additionally, as can be seen in Figures~\ref{fig:yy}D and~\ref{fig:shd}D, the test accuracy generally increases as more neurons are added to the network, which demonstrates an expected scalability with regard to the number of neurons. We can also see that training neurons' positions on top of synaptic weights (SpSNNs) outperforms training solely those weights (zero-dimensional networks).

For the YY task, the exact timing of the only spike that neurons emit carry information, which makes them quite sensitive compared to other coding schemes~\cite{bonilla2022ttfs}. Figure~\ref{fig:yy}D shows that SpSNNs can achieve up to 98.1$\pm$0.1\% test accuracy, which is on par with the results achieved in related work~\cite{goltz2025delgrad,goltz2021fast,Wunderlich2021eventprop}. This is despite the fact that SpSNNs are in a disadvantage in terms of temporal resolution due to their time-discretized simulations.
This shows that SpSNNs managed to learn the sensitive temporal requirements of the YY task, while maintaining their generality.

In contrast, the SHD task uses a rate-based encoding, which reduces, but does not eliminate, the importance of exact spike timing~\cite{guo2021coding}. We can see in Figure~\ref{fig:shd}D that SpSNNs achieve up to 83.6$\pm$0.9\% test accuracy with 300 recurrent neurons, a performance comparable to prior work~\cite{hammouamri2023learningdelays,cramer2020heidelberg,Mészáros2025efficient,sun2025attentional}. Notably, these previous approaches rely on additional mechanisms such as attention-based network architectures~\cite{sun2025attentional}, event-based simulation~\cite{Mészáros2025efficient}, or special measures during training~\cite{hammouamri2023learningdelays}, whereas SpSNNs achieve similar results within a more unified and flexible framework.

These results show that SpSNNs can capture complex temporal patterns, while displaying flexibility under different coding schemes, network architectures, and neuron spiking constraints. Additionally, different dimensionalities, including zero and infinite that refer to commonly seen SNNs, can easily be realized, learned, and tested in a single framework.

\subsection*{Regularization imposed by finite-dimensional SpSNNs improves accuracy}

So far, we observed that SpSNNs are able to learn to perform tasks of temporal nature by optimizing the positions of the neurons in multi-dimensional space instead of tuning the synaptic delay values of those neurons one by one. We next examine the way the choice of dimensionality influences SpSNN behaviour and performance. In doing so, we also compare SpSNNs against conventional SNN approaches.

If we assume an all-to-all recurrent SNN with $N$ neurons, explicitly training all synaptic delays means tuning $N^2$ parameters at each training step (excluding synaptic weights). The same recurrent neurons in a $D$-dimensional SpSNN require tuning $D\times N$ position parameters, with the large one-to-one delay matrix being the result of performing a mapping operation (Euclidean-distance calculation) on the smaller position matrix. Hence, in a mathematical sense, we can say that SpSNNs impose a regularization on the full one-to-one delay learning commonly found in delay-trainable SNNs. This regularization is similar to the Low-Rank Adaptation (LoRA)~\cite{hu2021lora} method used in the fine tuning of large language models. To better see the effect of this regularization imposed by SpSNNs, we can take a look at the test accuracy of the networks with regard to their number of tunable parameters. 

For the YY task, we can see that the two-dimensional SpSNN (which is the best-performing dimension) outperforms the infinite-dimensional network with the same number of parameters (Figure~\ref{fig:yy}E). This is also true for the SHD task, where both two- and three-dimensional SpSNNs outperform the network with infinite dimensions (Figure~\ref{fig:shd}E). These results show that the regularization that the SpSNNs provide in fact improves the test accuracy over the commonly seen networks with trainable delays for YY and SHD tasks. This can be better seen by plotting the accuracy of the networks against the dimension of the space for each number of hidden neurons. As shown in Figures~\ref{fig:yy}F and~\ref{fig:shd}F, the peak of the plot for almost all neuron counts is somewhere in the middle of the dimension spectrum (similar to hypothesis H2 in Figure~\ref{fig:setup}E), unlike our initial hypothesis where we assumed that a higher-dimension SpSNN would result in better performance (H1 in Figure~\ref{fig:setup}E).

This observation matches the results seen in other Artificial Neural Networks (ANNs) that employ similar regularization through low-rank factorization~\cite{hu2021lora,sainath2013lowrank}. The regularization that SpSNNs enforce seems to achieve better accuracy because it prevents memorization and guides the network towards a simpler solution. We performed further experiments to verify that the parameter reduction is indeed the source of the regularization imposed by SpSNNs, and not geometrical restrictions of the space. The most fundamental geometrical restriction in a Euclidean space with regard to distance is triangle inequality. We created SpSNNs that allow for non-straight connections between neurons (tortuous SpSNNs), effectively nullifying the triangle inequality (Figure~\ref{fig:sup3}A). We see that in these networks, the connections make use of the extra length they are allowed (Figure~\ref{fig:sup3}A), however, no visible changes in accuracy are inflicted by allowing such tortuousness (Figure~\ref{fig:sup3}B). Hence, we deduce that decreasing the parameter count is the source of the regularization enforced by SpSNNs.

Additionally, we observe that the best-performing dimension for each of our two tasks varies. For the YY task, the two-dimensional SpSNN performs the best, while for the SHD task the three-dimensional SpSNN outperforms the rest by a small margin. Though we cannot deduce general trends, it is nevertheless interesting that -- of all options in an infinite-dimensional space-- both tasks behave optimally for very low-dimensional configurations.

\begin{figure*}[t!]
    \centering
    \includegraphics[width=\textwidth]{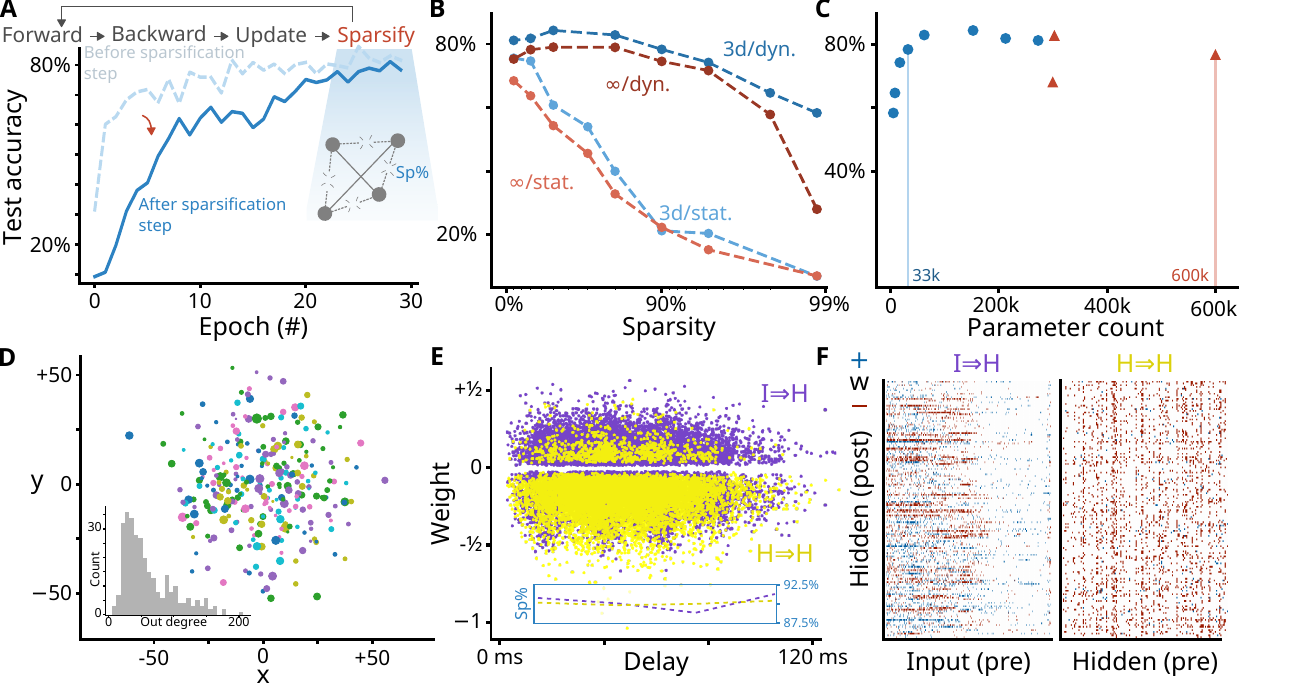}
    \caption{Sparse SpSNNs (Sp$^2$SNNs)
    \textbf{A:} Sp$^2$SNNs are built dynamically (dyn.) by adding a sparsification step during training after each epoch, where the weakest $Sp\%$ of connections are pruned. During earlier stages of the training, this extra step reduces accuracy as important connections are pruned, but as the network learns the right sparsity pattern, in the final epochs, Sp$^2$SNNs seem to be adapted to the pruning process and match the accuracy of the non-sparse network.
    \textbf{B:} Test accuracy versus sparsity rate for Sp$^2$SNNs and infinite-dimension networks. \textit{3d/stat} and \textit{$\infty$/stat} refer to post-training pruning strategies, while \textit{3d/dyn} and \textit{$\infty$/dyn} refer to the in-training, dynamic pruning described in pane A. For both networks, post-training pruning drops the accuracy significantly. On the other hand, dynamic pruning makes the network more resilient, with the Sp$^2$SNNs showing more robustness compared to infinite-dimensional SNNs.
    \textbf{C:} Test accuracy versus parameter count for different networks with 300-neuron networks. Blue points correspond to the different sparsity levels from B. The red points correspond to non-sparse networks. 90\% sparse Sp$^2$SNNs perform the same as infinite-dimensional network with 18$\times$ less parameters.
    \textbf{D:} The relation between position, degree of connection and graph clustering in the recurrent neurons. Different colors represent different Louvain communities. Node size corresponds to nodes degree, with its distribution shown in the inset.
    \textbf{E:} Synaptic delay vs synaptic distance. Each dot represents a retained synapse. Inset plot show relation between distance and sparsity per distance, measured in bins of 30ms. No strong relation seems to exists.
    \textbf{F:} Sp$^2$SNN weight matrix, showing positive (blue) and negative (red) weights. The network learns to ignore the high-frequency part of the input neurons.
    }
    \label{fig:sparse}
\end{figure*}

\subsection*{SpSNN accuracy is highly robust against sparsification}

Up to this point, we observed that SpSNNs are able to capture temporal patterns in different tasks, and in fact improve the test accuracy over infinite-dimensional networks while having less trainable parameters. To further reduce the parameter count of SpSNNs, we looked into pruning the synaptic weight of SpSNNs and making them sparse, thus introducing Sparse SpSNNs (Sp$^2$SNNs). Pruning has proven to be effective in reducing the number of parameters, and subsequently computational demand, for neural networks~\cite{hoefler2021sparsity} in general, and for SNNs specifically~\cite{han2023adaptivesparsestructuredevelopment,chakraborty2024sparsessn}. In addition to Sp$^2$SNNs, we studied the accuracy and behaviour of sparse infinite-dimensional networks and compared them to finite-dimensional ones.

As a first step, we statically pruned an already trained, non-sparse network by a sparsity ratio $Sp\%$. Here, we removed the weakest $Sp\%$ of the synaptic weights and benchmark the resulting network. The test accuracy results of the pruned three-dimensional SpSNNs and infinite-dimensional SNNs are shown in Figure~\ref{fig:sparse}B. We can see that following this pruning approach will result in a drop in test accuracy for both networks compared to their non-sparse counterparts; this holds true even for low sparsity rates of around $10\%$ and shows that these networks are susceptible to this post-training pruning strategy.

To overcome the challenges of the post-training sparsification, we employed a simple, dynamic-pruning strategy. In this approach, based on the sparsity ratio $Sp\%$ enforced, the weakest $Sp\%$ of the weights were set to zero after every epoch of training (Figure~\ref{fig:sparse}A). This strategy leads to a dynamically learned spatial structure (Figure~\ref{fig:sup2}). As can be seen in Figure~\ref{fig:sparse}A, during Sp$^2$SNN training, the test results are falling behind their non-sparse counterpart in the earlier stages. Towards the final training stages, the Sp$^2$SNN seems to be adapted to the pruning and matches the test accuracy of the non-sparse SpSNN. More importantly, in Figure~\ref{fig:sparse}B we can see that Sp$^2$SNNs can withstand sparsification of up to 90\% without a drop in accuracy. At the same time, unlike Sp$^2$SNNs, sparse, infinite-dimensional SNNs experience a drop in accuracy with increasing sparsity, as shown in Figure~\ref{fig:sparse}B. This means that, when looking at the test accuracy with regard to the parameter count (Figure~\ref{fig:sparse}C), we see that the 90\% Sp$^2$SNNs offer the same accuracy compared to the infinite-dimensional networks while having almost 18$\times$ less trainable parameters. In other words, finite-dimensional SpSNNs, when pruned with a dynamic, in-training strategy, are robust enough to withstand high levels of synaptic-weight sparsity while, at the same time, are more stable compared to infinite-dimensional networks despite having less parameters.

Furthermore, without explicitly enforcing it, Sp$^2$SNNs learn to ignore the input neurons associated with higher frequencies in the SHD task. This indicates that the network, through our pruning strategy, purposefully performs the sparsification process and selects the most important features of the input. We also observe that pruning in the Sp$^2$SNNs happens almost uniformly across connections regardless of their length, as can be seen in Figure~\ref{fig:sparse}E. This outcome goes against our intuition -- drawn from our observations of brain functionality~\cite{Chklovskii2002wiring,Kolodkin2010wiring} -- that Sp$^2$SNNs should favour shorter connections due to their immediate effect on the post-synaptic neurons. This is, then, further corroborative evidence that striving to mimic the brain should not interfere with technical solutions which are effective, even if departing from biorealism.

\subsection*{General software framework for implementing SpSNNs}
Thus far, the effectiveness and performance of SpSNNs have been demonstrated. This new class of SNNs not only provides a unified framework that encompasses existing architectures but also exposes new dimensionalities for exploration. It has been shown that finite-dimensional networks can outperform their infinite counterparts and exhibit remarkable robustness to sparsification. Because SpSNNs constitute a general framework that imposes no constraints on the underlying network, they were implemented using an \textit{equally} general, novel software infrastructure.

The SpSNNs were configured differently for the two tasks considered in this work, as each required distinct network architectures, encoding schemes, spiking constraints, and neuron types (see Figure~\ref{fig:adex}). This level of implementation flexibility was enabled by putting together a universal software framework for SpSNNs. Our key insight at this point was that gradients must not be computed manually. Hand-derived, exact gradients in this domain are very task- and model-specific: the expressions and tractability of analytical solutions change depending on different aspects of the underlying network.

This motivates (1) a main implementation requirement for automatic differentiation (autograd). We employed JAX~\cite{jax2018github,jax2018compiling} which provides a highly composable autograd system. However, such systems, in their basic implementations, are unable to differentiate SNNs correctly due to the spike-induced voltage-reset mechanism inherent in SNN simulations and also due to ignoring delay gradient during the backward pass, in networks with trainable synaptic delays.

To overcome these two challenges, we have to guide JAX's autograd to correctly calculate the exact gradients by carefully handling the discontinuities caused by the reset mechanism, in addition to preserving the delay-parameter gradients; the latter is addressed by our previous work, EventQueues~\cite{landsmeer2025eventqueues}. We address the former challenge as part of this work by annotating spike-dependent expressions with custom gradients during the SNN simulation: In the backward pass, spike-timing information is inserted during spike-induced state-variable updates, and gradients are retained through voltages resets (see Methods). This approach, which to the best of our knowledge is the first to automatically derive exact gradients, not only applies to SpSNNs but also to SNNs in general, specially ones with trainable delays.

Additionally, for the purpose of implementation generality, (2) we employed time-discretized simulations in this work instead of event-based ones, as the latter would limit different aspects of the network such as type of the neurons~\cite{Brette2007simulation} and their parameters~\cite{goltz2021fast} in addition to possible network architectures~\cite{engelken2023sparseprop}. This is due to the fact that event-based simulators work best with simpler, analytically solvable neuron models. On top of that, in densely connected, highly recurrent networks, an event-based simulation loses its performance advantages over a time-driven one due to the large number of recurrent spike events. Hence, the flexibility of our developed method for SpSNNs makes it easy to test and deploy different spatial networks, without having to manually derive the adjoint dynamics (see Methods).

\section*{Discussion}

SNNs, at large, aim to achieve high energy efficiency and intelligence by mimicking the brain. Here, we introduced Spatial Spiking Neural Networks (SpSNNs), a new class of networks in which neurons occupy a Euclidean space with learnable positions, and communication delays between neurons arise from their distance in this space. By optimizing neuron positions rather than individual delay values, SpSNNs substantially reduce the number of trainable parameters while retaining the benefits of delay-based temporal expressiveness~\cite{hammouamri2023learningdelays,goltz2025delgrad,sun2024delay,sun2023learnable,Mészáros2025efficient,queant2025delrec}. When sparse, SpSNNs provide great robustness, order-of-magnitude parameter saving without sacrificing accuracy, and improved scalability, particularly for recurrent networks. Additionally, the novel autograd-based implementation we developed for this work not only enables the scalable and frictionless training of SpSNNs, but SNNs in general.

We view \textit{efficient machine intelligence} as a combination of (1) a high ratio of a model's accuracy to its computational effort in addition to (2) compatibility of those computations with the underlying hardware substrate. Based on our findings, we argue that SpSNNs increase intelligence efficiency by improving upon both of these factors. By achieving higher accuracy, while having a lower number of parameters, and as a result less computation, SpSNNs offer a path to higher levels of machine-intelligence density with regard to computational effort. Additionally, since the spatial structure of SpSNNs can directly translate into geometrical placements, it in principle allows for efficient mapping to 2D and 3D chip structures. This comes in stark contrast to networks with unconstrained delays, where the extra parameters would lead to higher computational demand, and the unboundedness of the delay values would limit the performance and programmability of the underlying hardware.

Several research directions emerge from this work. First, due to the generality of our implementation, SpSNNs could be evaluated under event-driven simulation and extended to more biophysically realistic neuron and synapse models (see Figure~\ref{fig:adex}), including processes such as myelination dynamics~\cite{waxman1980myelin}. The latter could open the door for the SpSNN framework to be used for neuroscientific studies. Second, translating the spatial abstraction of SpSNNs into hardware requires investigating both analogue and digital neuromorphic substrates that can exploit position-aware computation. Third, the relationship between task dimensionality and the optimal spatial dimensionality of SpSNNs remains unexplored in this work and warrants systematic benchmarking across diverse tasks. Together, these avenues offer opportunities to strengthen the biological grounding, computational efficiency, and hardware readiness of SpSNNs.

Despite the presented SpSNN novelties, some limitations exist within this work. Firstly, we chose a time-discretized simulation approach with a coarse-grained timestep. For truly exact gradients, an event-based simulation strategy could replace our time-discretized transient simulation. However, the same custom gradients developed as part of this work can still be used for event-based simulation. Secondly, we make typical assumptions about the underlying neural models: a first-order synaptic current, no instantaneous voltage changes when receiving a spike, and linear transmission from the current to the membrane voltage. Of course, more complex, biorealistic models have different equations, potentially requiring a different custom gradient and violating current autograd functionality. Still, the methodology for deriving new custom gradients, as presented in this work, would apply to any further refinements.

The promising results exhibited by SpSNNs in this work show that brain-inspired ideas and mechanisms provide a path towards increasingly efficient intelligence. More broadly, certain ANN architectures could potentially benefit from the core idea of SpSNNs -- e.g., Neural Delay Differential Equations~\cite{zhu2021neural}) -- by incorporating our methodology for trainable delays or for bringing these into the spatial domain. Overall, we believe that the ideas presented in this work and the SpSNN framework can unlock ultra-efficient and capable intelligent systems across both traditional AI infrastructures and emerging analogue and digital neuromorphic chips.

\section*{Methods}

\subsection*{Mapping neurons to a space}

In principle, we can map neurons to any other non-Euclidean spaces as well and the SpSNNs, in addition to their implementation, do not prevent us from doing so. However, Euclidean space is of special interest for us as humans live in one and also we care about realization and manufacturability of SpSNNs on hardware. Additionally, we note that in Euclidean spaces, certain geometric properties that are based on norms, such as the triangle inequality, extend from finite-dimensional Euclidean spaces to an infinite-dimensional space. Hence, referring to a network where synaptic delays can take any values without constraint as "infinite-dimensional" is not strictly accurate, but it is sufficiently precise for our purposes and does not affect the argument.

\subsection*{Neuron models}

For the Leaky Integrate-and-Fire (LIF) neuron cells used we have the following dynamics.

\begin{align}
\dot i_{\rm syn} &= -\frac1{\tau_{\rm syn}} i_{\rm syn} + \sum w {\delta(t - t_{\rm post})} \label{eq:doti}
\\
\dot v &= -\frac1{\tau_{\rm mem}} v + i_{\rm syn}
\label{eq:dotv}
\end{align}

For AdEX neurons, we keep the same synaptic current, but change the voltage equation and add an extra adaptation current

\begin{align}
\dot v &= -\frac1{\tau_{mem}} \left(
v + \Delta_{T} \exp\left(
\frac{v-\frac12}{\Delta_{T}}
\right)
\right) + i_{\rm syn} - i_{\rm adapt}
\\
\dot i_{adapt} &=
\frac1{\tau_{\rm adapt}} \left(
- i_{\rm adapt} + a v
\right)
+
b \sum {\delta(t - t_{\rm pre})}
\label{eq:iadapt1}
\end{align}

In both cases, reset voltage is set at 0 and the threshold voltage is 1.

\subsection*{Event-aware time-discretization}

To enable trainable delays in an autograd framework, we need to solve
two problems: delays and resets.
Delays only occur as array-indices in a time-discretized implementation and are thus not seen by the autograd framework, hence requiring custom gradients to retain their effects.
Resets in neural models, in a time-discretized frame, reset the voltage to zero, which also zeros out the gradient.

To overcome these problems, we need to temporarily switch to an event-aware frame, where we carefully lift over gradients from before (denoted with a minus sign, -) to after (denoted with a plus sign, +) the generation and reception of spikes.
To this extent, we adopted delay gradients and the implementation of spike-queues from EventQueues~\cite{landsmeer2025eventqueues}.
To solve reset gradients, we generalized some of the methods used in  EventProp~\cite{Wunderlich2021eventprop}.
For ease of implementation, custom gradients are only defined using the Jacobian-Vector Product (JVP), and JAX is used for deriving the Vector-Jacobian Product (VJP) actually used in the backward pass.

We time-discretize the neuron equations using a fixed timestep. In this frame, the voltage resets to zero when the neuron spikes.
When a neuron spikes, the voltage-reset is handled using the v\_reset function, which has a custom gradient as defined in equation~\ref{tvnext}. The v\_reset function either resets the voltage to zero, when the neuron spikes, or to $v_{next}^{noreset}$ when it doesn't spike. For correctly handling the gradients, and not just resetting them to zero as well, we need the values of the voltage-time gradients, as well as the tangents of the voltage, which we provide as extra arguments to the function. At each timestep, the next values for voltage and current are calculated as equations~\ref{eq:vnext1} and~\ref{eq:inext1}.
In general, we use forward Euler discretization, but where possible, we employ the exact analytical exponential solution for LIF cells.

\begin{align}
S &=
\begin{cases}
1 & \text{if } v \ge 1 \\[2pt]
0 & \text{otherwise}
\end{cases} \\[2pt]
v_{\rm next} &= \text{v\_reset}\big(S, v,
\left[ \dot v^- \right],
\left[ \dot v^+ \right],
v^{\rm noreset}_{\rm next}\big)\label{eq:vnext1}
\\
i_{\rm next} &= \exp\left(-\frac{\Delta t}{\tau_{syn}}\right)\cdot i_{\rm syn} + i_{\rm syn}^{\rm jump}\label{eq:inext1}
\end{align}

Importantly, the state variables $v_{\rm next}$ and $i_{\rm next}$ depend on an additive variable $v_{\rm jump}$ or $i_{\rm jump}$ which codifies the effect of spikes. In the next two sections, we will describe these variables, and importantly, their gradients.
As mentioned earlier, spikes in transit are stored in EventQueues~\cite{landsmeer2025eventqueues}. For GPU-based optimal simulation with coalesced memory-access, and because the additive effects of spikes are linear, we use a single ring-buffer per neuron. The ring-buffer is modified from the original work to store both $v_{\rm jump}$ or $i_{\rm jump}$, instead of just $i_{\rm jump}$. Note that $v_{\rm jump}$ is necessarily just a array of zeros in the queue, but it will carry over the associated gradients from presynaptic neurons in the tangent space.

For LIF cells, the arguments to the v\_reset() function in equation~\ref{eq:vnext1} become:

\begin{align}
v^{\rm noreset}_{\rm next} &=
\exp\Big(-\frac{\Delta t}{\tau_{\rm mem}}\Big)
\cdot v + I_{\rm syn} \cdot \Delta t + v_{\rm jump} \label{vnoreset} \\[4pt]
\left[ \dot v^- \right]&= I_{\rm syn} - \tau_{\rm mem} \cdot v \label{dvm} \\[4pt]
\left[ \dot v^+ \right]&= I_{\rm syn} + i_{\rm jump} \label{dvp}
\end{align}

For AdEx neurons, the method for imposing timing-aware derivatives on the voltage equation is the same. However, the arguments to the v\_reset() function (equations \ref{vnoreset}, \ref{dvm} and \ref{dvp}) change to the corresponding AdEx model versions.
On top of that, we also integrate the current $i_{\rm adapt}$ using forward Euler integration alongside equations~\ref{eq:vnext1} and~\ref{eq:inext1}, using a similar $i^{\rm jump}_{\rm adapt}$ to handle the discontinuities (discussed next).

\subsection*{Delay gradients}


To enable autograd-based trainable delays,
instead of manually deriving adjoint dynamics, we assume changes in the time-discretized voltages, synaptic and adaptation currents to come in the form of $v_{jump}$, $i_{jump}$ and $i_{adapt}^{jump}$ respectively.
In primal space, $v_{jump}$ is zero, $i_{jump}$ is the synaptic weight and $i_{adapt}^{jump}$ is the adaptation constant $b$.
By annotating the tangent space representations of these values with the right derivatives, we can handle the spike-time and delay gradients.
We do this via a custom gradients, defined via a custom JVP. Here, we denote the semantics of these gradients by explicitly writing out the tangent space equations, using the operator
$\mathrm T\left[
\cdot
\right]
$
for tangent space.
For simplicity, we start with writing the JVPs here as used in the implementation, before showing the derivation in the next section.

The derivative of spike times follows the following well-known equation~\cite{yang2014proof}, which we have amended here with a delay gradient. In the implementation, equation~\ref{tpostt} is substituted directly in the calculation of the $v_{\rm post}^{\rm jump}$ and $i_{\rm syn}^{\rm jump}$ gradients, this equation~\ref{tpost} is not actually used there. $t_{\rm simulation}$ just refers to the current time in the simulation. For the TTFS scheme, final spikes times are outputted using equations~\ref{tpost} and~\ref{tpostt} to correctly handle the loss-gradients.

\begin{align}
t_{\rm post}
&\coloneq t_{\rm simulation} + 
\mathrm{delay} \label{tpost}
\\
\mathrm T\left[
t_{\rm post}
\right]
&\coloneq -\,\frac{
\mathrm T\left[
v_{\rm pre}
\right]
}{\dot v_{\rm pre}} + 
\mathrm T\left[\mathrm{delay}\right]\label{tpostt}
\end{align}

For a recurrent connection (between two spiking neurons), we have the following for the synaptic current jump and the tangent of the synaptic current jump:

\begin{align}
i^{\rm jump}_{\rm syn} &\coloneq w \\[6pt]
\mathrm T\left[
i^{\rm jump}_{\rm syn}
\right]
&\coloneq 
\mathrm T\left[
w_t
\right]
+ \frac{w}{\tau_{\rm syn}}\,
\mathrm T\left[
t_{\rm post}
\right]
\end{align}

For the voltage jump and the tangent of the voltage jump we have:

\begin{align}
v^{\rm jump}_{\rm post} &\coloneq  0 \\[6pt]
\mathrm T\left[
v^{\rm jump}_{{\rm post}}
\right]
&\coloneq 
w
\, 
\mathrm T\left[
t_{\rm post}
\right]\label{vjumpt}
\end{align}

For the adaptation current, if we interpret the spike as an instantaneous increase in adaptation current $i_{\rm adapt}^{\rm jump}$, we have:

\begin{align}
i_{\rm adapt}^{\rm jump} &= b
\\
T\left[
i_{\rm adapt}^{\rm jump}
\right]
& =
\frac{b}{\tau_{\rm adapt}}
T\left[
t_{\rm pre}
\right]
\end{align}

For input neurons, which have static spike times, equation~\ref{tpostt} reduces to:

\begin{align}
\mathrm T\left[
t_{\rm post}
\right]
&= 
\mathrm T\left[\mathrm{delay}\right]
\end{align}

after which we use JAX's auto-differentiation to derive the VJP, which indicates how the output's gradient propagates back to the inputs.

\subsection*{Delay gradient derivation}

The previously presented delay gradients were derived inspired by methods presented in EventProp~\cite{Wunderlich2021eventprop}. In EventProp, the gradients towards weight parameters were manually derived.
To create custom gradients, as JVPs, fitting in an autograd framework, these methods need to be generalized towards an arbitrary parameter $x$.

We shortly repeat our derivations here. To obtain JVPs, we write the relation between the total derivatives of the each state variable towards an arbitrary variable $x$, before (-) and after (+) the arrival of the spike. Importantly, these total derivatives contain a partial derivative through the spike-time dependence, which we split out as the product between the state-time gradient and the spike-time gradient. Substituting in the actual state-time gradient before and after the spike-event, and keeping the spike-time gradient as is, using equation~\ref{tpostt}, we obtain our JVP as the difference of the partial derivative of the state towards $x$ before and after the spike.

\begin{itemize}
\item
For the JVP of $v_{\rm jump}$, on receiving a spike,
we have  $v^+ = v^-$, which means the $v_{\rm jump} = 0$.
However, differentiating this on both sides gives us

\begin{align}
 \frac{\partial v^+}{\partial x} +
 \frac{\partial v^+}{\partial t_{\rm post}}
 \frac{\partial t_{\rm post}}{\partial x}
 &=
 \frac{\partial v^-}{\partial x} +
 \frac{\partial v^-}{\partial t_{\rm post}}
 \frac{\partial t_{\rm post}}{\partial x}
\end{align}

which we can rewrite as

\begin{align}
 \frac{\partial v^+}{\partial x} 
 &=
 \frac{\partial v^-}{\partial x} +
 \left(
 \frac{\partial v^-}{\partial t_{\rm post}}
 -
 \frac{\partial v^+}{\partial t_{\rm post}}
 \right)
 \frac{\partial t_{\rm post}}{\partial x}\label{eq:nice}
\end{align}

The time-derivative of the voltage does change because the synaptic current jumps with $w$.

\begin{align}
 \dot v^+ &=
 \dot v^- + w
\end{align}

which we can input back in equation~\ref{eq:nice} as

\begin{align}
 \frac{\partial v^+}{\partial x}
 &=
 \frac{\partial v^-}{\partial x} -
w
 \frac{\partial t_{\rm post}}{\partial x}
\end{align}

\item
For the JVP of $i_{\rm jump}$, we have

\begin{align}
i_{\rm syn}^+ = i_{\rm syn}^- + w
\label{ijumporig}
\end{align}

which we can differentiate, then rewrite, to

\begin{align}
\frac{\partial i_{\rm syn}^+}{\partial x}
=
\frac{\partial i_{\rm syn}^-}{\partial x} +
\left(
\frac{\partial i_{\rm syn}^-}
{\partial t_{\rm post}}
-
\frac{\partial i_{\rm syn}^+}{\partial t_{\rm post}}
\right)
\frac{\partial t_{\rm post}}{\partial x}
+  \frac{\partial w}{\partial x}
\label{inice}
\end{align}

Then, combining equations~\ref{ijumporig} and~\ref{eq:doti}

\begin{align}
\dot i_{\rm syn}^+ = \dot i_{\rm syn}^- - \frac{w}{\tau_{\rm syn}}
\end{align}

which we can substitute back in equation~\ref{inice}, obtaining

\begin{align}
\frac{\partial i_{\rm syn}^+}{\partial x}
=
\frac{\partial i_{\rm syn}^-}{\partial x}
- \frac{w}{t_{\rm \rm syn}}
\frac{\partial t_{\rm post}}{x}
+  \frac{\partial w}{x}
\end{align}

\item
For the JVP of $i_{\rm adapt}$, we have, starting from the update
\begin{align}
 i_{\rm adapt}^+ &=
 i_{\rm adapt}^- + b
\end{align}

We can differentiate both sides, assuming $b$ to be constant, to obtain

\begin{align}
\frac{\partial i_{\rm adapt}^+}{\partial x}
+
\dot i_{\rm adapt}^+
\frac{\partial t_{\rm pre}}{\partial x}
&=
\frac{\partial i_{\rm adapt}^-}{\partial x}
+
\dot i_{\rm adapt}^-
\frac{\partial t_{\rm pre}}{\partial x}
\end{align}

Then substituting in equation~\ref{eq:iadapt1} and rewriting a bit, we obtain our JVP:

\begin{align}
\\
\frac{\partial i_{\rm adapt}^+}{\partial x}
& =
\frac{\partial i_{\rm adapt}^-}{\partial x}
+
\frac{b}{\tau_{\rm adapt}}
\frac{\partial t_{\rm pre}}{\partial x}
\end{align}

\end{itemize}

\subsection*{Reset gradients}

As stated earlier, correct handling of gradients around the reset discontinuity is done using a v\_reset() function, repeated here for clarity:

\begin{equation}
v_{\rm next} =
\text{v\_reset}\big(S, v, \left[ \dot v^- \right], \left[ \dot v^+ \right], v^{\rm noreset}_{\rm next}\big)
\end{equation}

It requires the exact voltage time-derivatives before and after the reset;
$\left[ \dot v^- \right]$ and $\left[ \dot v^+ \right]$.
These only need to be calculated for correctly handling spikes, not to actually update the voltage. We denote fact this with the square brackets.

Note that in equation~\ref{vnoreset}, the $v_{jump}$ is zero, but does carry a tangent over from the previous definition in equation~\ref{vjumpt}  through autograd.
$v\_reset()$ is a function with a custom JVP.
For given primal inputs
$S, v,
\left[ \dot v^- \right],
\left[ \dot v^+ \right],
v^{\rm noreset}_{\rm next}$ and tangents $\mathrm T\left[v_t\right]$ and $
\mathrm T\left[v^{\rm noreset}_{\rm next}\right]$, it returns the next voltage primal $v_{\rm next}$ and tangent $\mathrm T\left[ v_{\rm next}\right]$:

\begin{align}
v_{\rm next}
&=
\begin{cases}
0 & \text{if } S = 1 \\[2pt]
v^{\rm noreset}_{\rm next} & \text{otherwise}
\end{cases}
\\[2pt]
\mathrm T \left[v_{\rm next}\right] &=
\begin{cases}
{\displaystyle
\frac{
\left[ \dot v^+ \right]
}{
\left[ \dot v^- \right]
}
}
\, \mathrm T\left[v\right]
& \text{if } S = 1 \\[2pt]
\mathrm T\left[ v^{\rm noreset}_{\rm next}\right] & \text{otherwise}
\label{tvnext}
\end{cases}
\end{align}

Again, equation \ref{tvnext} derives as a generalization over EventProp. Starting from the reset point $v^+ = 0$, we obtain by differentiation

\begin{align}
    \frac{\partial v^+}{\partial x} +
    \frac{\partial v^+}{\partial t_{\rm pre}}
    \frac{\partial t_{\rm pre}}{\partial x} = 0
\end{align}

If we then substitute in the spike-time gradient~\cite{yang2014proof} from equation~\ref{tpostt}, we end up with

\begin{align}
    \frac{\partial v^+}{\partial x} = 
    \frac{\dot v^+}{\dot v^-}
    \frac{\partial v^-}{\partial x}
\end{align}

\subsection*{SpSNNs}

Using the defined methods, we can now implement SpSNNs using regular autograd, using JAX to propagate gradients through the delay calculations. We simply state here the primal equations.

As mentioned earlier, SpSNNs see neurons in a Euclidean space. Hence, for each input or hidden neuron $i$, a trainable position vector $\vec r_i$ is defined. The communication delays between neurons are calculated based on their distance. These delays are calculated as:

\begin{align}
    d_{ij} = \left|\vec r_i - \vec r_j\right|_2
\end{align}

For Tortuous SpSNNs, additional variation via a trainable parameter $E_ij$ around the distance is allowed as mentioned earlier. Delay calculation in these network are as:

\begin{align}
d_{ij} = \frac12 \left(1 + \epsilon \tanh(E_{ij})\right) \left|\vec r_i - \vec r_j\right|_2
\end{align}

Here, $\epsilon$ denotes the maximum deviation from the straight line distance.

We showed the JVP of the equations based on the delay $d_{ij}$ earlier. The position of each neuron then is optimized by using these equation using autograd.

\subsection*{TTFS Network}
For the TTFS encoding used for the YY task, each neuron was limited to only spike once. The output neurons were encoded with TTFS, where an earlier spike indicates the classification of the network. Hence, when training, we want the correct neuron to spike first. To emphasize this during training, we used a hinge loss function (equation~\ref{eq:hinge}) mostly common in Support-Vector Machines (SVM)~\cite{Cortes1995svm}.

\begin{align}
    &\mathcal{L} = \sum_{k\neq correct} \mathrm{softplus}(\beta(t_{correct} - t_{k} + m)) \label{eq:hinge} \\
    &\mathrm{softplus}(z) = \log (1+e^z)
\end{align}

$t_{correct}$ is the spike time of the correct neuron, $t_k$ is the spike time of all the other neurons, $\beta$ is the slope of the loss function, and $m$ is the desired margin we want to have between the spike times of correct and other neurons.

\subsection*{Rate-coded Network}

SpSNNs were configured with a rate-coded scheme for the SHD task. A linear readout of the spikes was used in the these SpSNNs. For these networks, we used the superspike formalism~\cite{zenke2018superspike}.
Superspike for the rate-coded output was found to make training easier in the absence of spikes.

\begin{align}
    S_t^{j} &= \tilde H(V_t^{j}-v_{th})
    \\
    \tilde H(x) &= 
    \begin{cases}
        1 & x \ge 0 \\
        0 & x < 0 \\
    \end{cases}
    \\
    \quad
    \frac{\partial \tilde H(x)}
         {\partial x} &= \frac{1}{\left(\left|x\right|+1\right)^2}
         \label{eq:superspike}
    \quad
\end{align}

For the linear readout, a matrix $W_{\rm readout} \in \mathbb{R}^{N_{o} \times N_{h}}$ is also trained.

\begin{align}
p_i = \frac{e^{\left(WS_t\right)_i}}{\sum_j e^{\left(WS_t\right)_j}}
\\
Y_i =  \mathrm{argmax}_i \, z_i
\left(WS_t\right)_i
\end{align}

Cross entropy loss function:

\begin{align}
\mathcal{L}_{\text{CE}}(p, y)
    &= - \sum_{i} y_i \,\log p_i
\end{align}

\subsection*{Training}

For both tasks, datasets were stored in \texttt{.hdf5} format. Input data were loaded into the GPU memory, with spikes being represented as bit-packed 32-bit integers, and cached to optimize data throughput during training. Additionally, to reduce memory footprint during simulations, we performed checkpointing every 100 timesteps. All models were, unless otherwise noted, trained using the Adam optimizer. Initial developments and debugging were performed on Nvidia RTX Pro 6000 and Quadro RTX 6000 GPUs. The training of all models was performed on NVIDIA H100 GPUs.

\subsubsection*{YY Task}
Table~\ref{tab:yy_training} summarizes the hyper-parameters associated with SpSNNs' training for the YY task. For this task, because of the sensitive TTFS encoding used, the learning rate was tuned specifically for each set of hyper-parameters.

\begin{table}[t]
    \centering
    \begin{tabular}{c|c}
      \textbf{Hyper-parameter} & \textbf{Value} \\ \hline
        Training Epochs & 300 \\
        Batch Size & 150 \\
        Learning Rate (LR) & $5 \times 10^{-5}$ - $1 \times 10^{-3}$ \\
        LR Scheduler & Cosine-decay \\ 
        LR Schr. Steps  & 10,000 steps \\
        LR Schr. Warm-up & 500 steps \\ 
        LR Schr. Final & 10\% \\
        Scale Factor & 1.0\\
        Adam & (0.9, 0.999, 10$^{-8}$)\\
    \end{tabular}
    \caption{Hyper-parameter used for training SpSNNs for the YY task.}
    \label{tab:yy_training}
\end{table}

\begin{table}[t!]
    \centering
    \begin{tabular}{c|c}
      \textbf{Hyper-parameter} & \textbf{Value} \\ \hline
        Training Epochs & 30 \\
        Batch Size & 32 \\
        Learning Rate & $1 \times 10^{-3}$ \\
        LR Scheduler & Cosine-decay \\ 
        LR Schr. Steps  & 10,000 steps \\
        LR Schr. Warm-up & 500 steps \\ 
        LR Schr. Final & 10\% \\
        Scale Factor & 10.0 \\
        Adam & (0.9, 0.999, 10$^{-8}$)\\
    \end{tabular}
    \caption{Hyper-parameter used for training SpSNNs for the SHD task.}
    \label{tab:shd_training}
\end{table}

\subsubsection*{SHD Task}
Table~\ref{tab:shd_training} summarizes the hyper-parameters used for training SpSNNs for the SHD task. Input features representing position and delay were scaled by a factor of 10 to improve numerical stability during training.

\clearpage
\clearpage

\paragraph{Acknowledgments}
This paper is partially supported by the European-Union Horizon Europe R\&I program through projects SEPTON (no. 101094901) and SECURED (no. 101095717) and through the NWO - Gravitation Programme DBI2 (no. 024.005.022). This work used the Dutch national e-infrastructure with the support of the SURF Cooperative using grant no. EINF-10677, EINF-15815, and EINF-16791. The RTX PRO 6000 and Quadro Pro 6000 GPUs used for this research were donated by the NVIDIA Corporation.

\paragraph{Code Availability}
We plan to make the SpSNN framework publicly available for researchers and interested parties.

\bibliographystyle{unsrt}
\bibliography{main}

\onecolumn
\renewcommand{\thefigure}{S\arabic{figure}}
\setcounter{figure}{0}
\clearpage

\section*{Supplementary Information}

\begin{figure*}[h]
    \centering
    \includegraphics[width=0.9\textwidth]{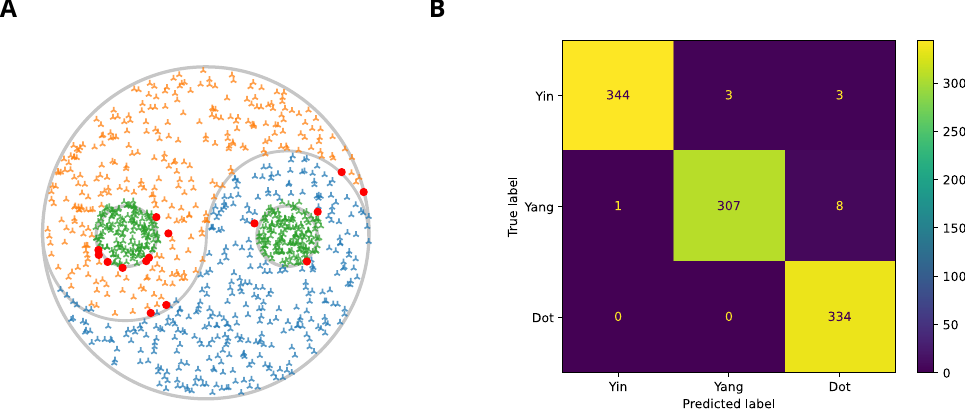}
    \caption{YY classification statistics
    \textbf{A)} Misclassified points. We see that the most misclassified points are along the borders of the regions.
    \textbf{B)} Confusion matrix
    }
    \label{fig:sup1}
\end{figure*}

\begin{figure*}[h]
    \centering
    \includegraphics[width=0.8\textwidth]{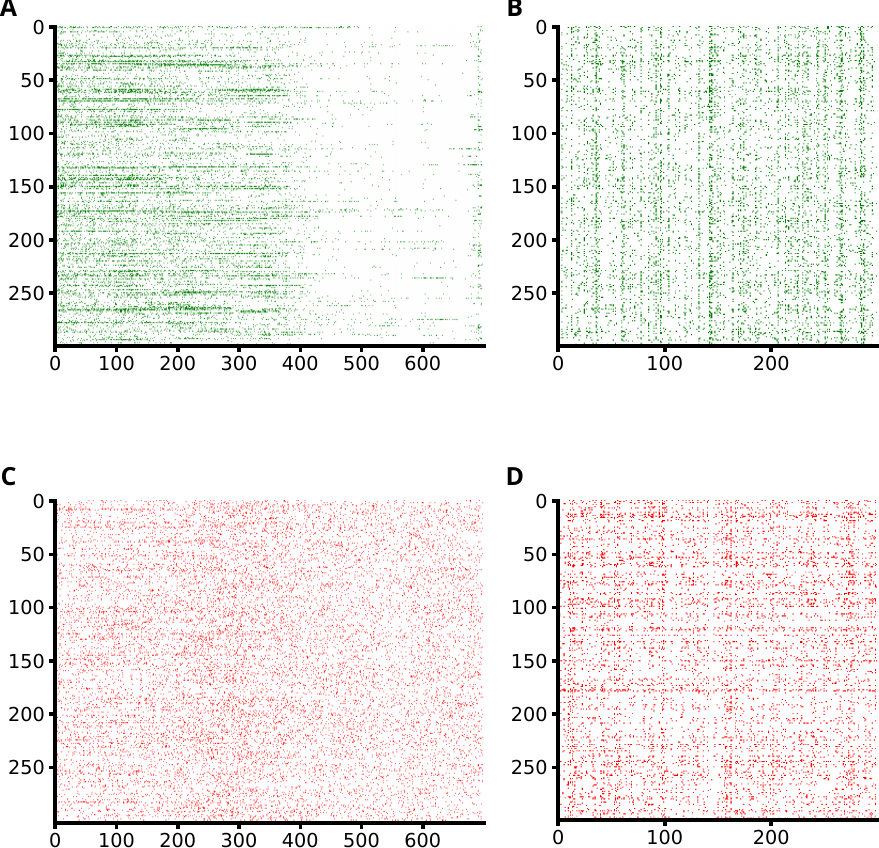}
    \caption{
    Changes induced by dynamic sparsity between epoch 1 and epoch 30 for the SHD task
    \textbf{A)} Created input to hidden connections
    \textbf{B)} Created hidden to hidden connections
    \textbf{C)} Removed input to hidden connections
    \textbf{D)} Removed hidden to hidden connections
    }
    \label{fig:sup2}
\end{figure*}

\begin{figure*}[h]
    \centering
    \includegraphics[width=0.6\textwidth]{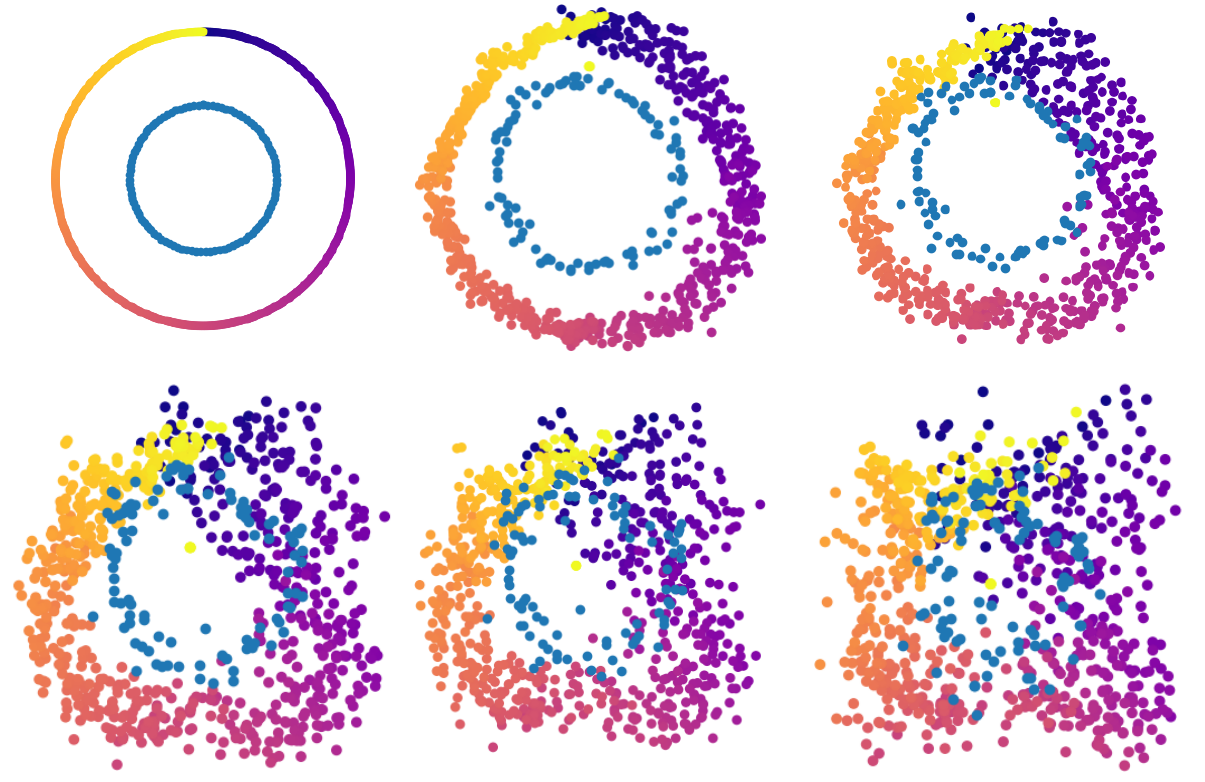}
    \caption{
    Position changes during training, when initialized from two circles and all initial weights equal, showing that changes are not only local.
    }
    \label{fig:sup3}
\end{figure*}

\begin{figure*}[h]
    \centering
    \includegraphics[width=0.6\textwidth]{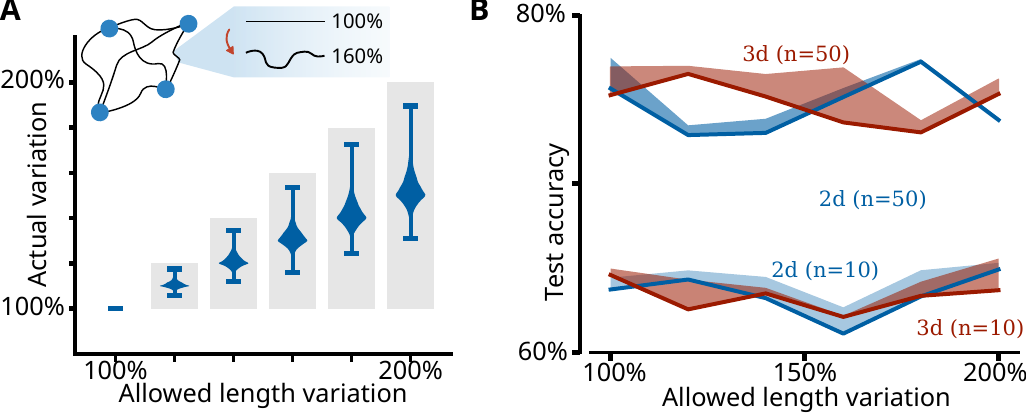}
    \caption{Tortuous SpSNNs
    \textbf{A:}
    Tortuous SpSNNs allow a biorealistic bounded extra length variation, on top of the direct euclidean distance. This extra distance is actually used during training.
    \textbf{B:}
    This extra variation seems not to have significant effect of performance.
    }
    \label{fig:sup3}
\end{figure*}

\begin{figure*}[h]
    \centering
    \includegraphics[width=\textwidth]{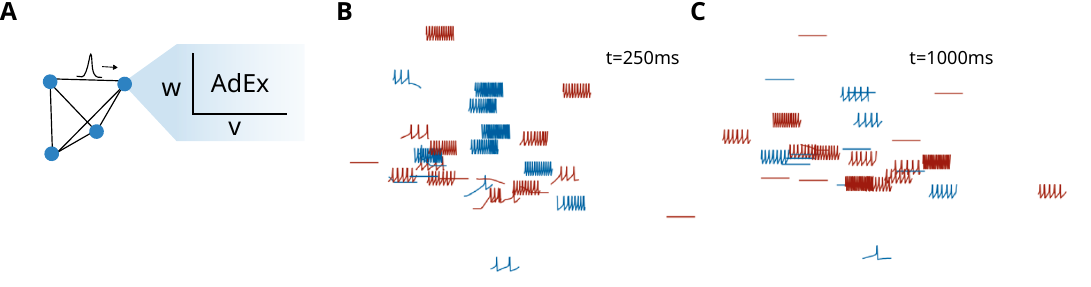}
    \caption{SpSNNs with AdEx neurons
    \textbf{A:} AdEx~\cite{brette2005adex} neuron model is used
    \textbf{B/C:} Network seem to work in two stages of receiving input and outputting result.
    Red neurons are the neurons with the highest sum-of-absolute weights in the readout matrix, while blue neurons are the least output-associated.
    }
    \label{fig:adex}
\end{figure*}

\end{document}